\documentclass[runningheads]{llncs}
\usepackage{graphicx}
\usepackage{acro}
\usepackage{algorithm}
\usepackage{algorithmic}
\usepackage{subfigure}
\usepackage{caption}
\usepackage{epsfig}
\usepackage{epstopdf}
\usepackage{wrapfig}
\usepackage{multirow}
\usepackage{array}
\usepackage{tabulary}
\usepackage{graphicx}
\usepackage{amsmath}
\usepackage{amssymb}
\usepackage{mwe}
\usepackage{amsmath,amssymb}
\usepackage{booktabs}
\DeclareAcronym{CT}{
short=CT,
long=computed tomography,
long-plural-form = computed tomographies,
}

\DeclareAcronym{MRI}{
short=MRI,
long=magnetic resonance imaging
}

\DeclareAcronym{TLV}{
short=TLV,
long=total liver volume
}
\DeclareAcronym{FRLV}{
short=FRLV,
long=future remnant liver volume
}

\DeclareAcronym{LDLT}{
short=LDLT,
long=living donor liver transplantation
}

\DeclareAcronym{CAST}{
short=CAST,
long=computer assisted segmentation tool
}

\DeclareAcronym{DEXTR}{
short=DEXTR,
long=deep extreme points
}

\DeclareAcronym{ADA}{
short=PADA,
long=prediction-based adversarial domain adaptation
}

\DeclareAcronym{PHNN}{
short=PHNN,
long=progressive holistically nested network
}

\DeclareAcronym{method}{
short=UGDA,
long=user-guided domain adaptation
}

\DeclareAcronym{RW}{
short=RW,
long=random walker
}

\DeclareAcronym{CRF}{
short=CRF,
long=conditional random field
}
\DeclareAcronym{mm}{
short=mm,
long=millimeter
}

\DeclareAcronym{FCN}{
short=FCN,
long=fully-convolutional network
}

\DeclareAcronym{DSC}{
short=DSC,
long=Dice-S\o{}rensen coefficient,
}
\DeclareAcronym{PCK}{
short=PCK,
long=Percentage of correct Key-points,
}
\DeclareAcronym{PACS}{
short=PACS,
long=picture archiving and communication system,
}
\DeclareAcronym{MXA}{
short=MXA,
long=mask-extreme-point agreement,
}

\DeclareAcronym{ICC}{
short=ICC,
long=intrahepatic cholangiocellular carcinoma
}

\DeclareAcronym{CGMH}{
short=CGMH,
long=Chang Gung Memorial hospital
}

\DeclareAcronym{HCC}{
short=HCC,
long=hepatocellular carcinoma
}

\DeclareAcronym{PS}{
short=UI,
long=user interaction,
}

\newcommand{\etal}{\textit{et al.}}
\newcommand{\eg}{\textit{e.g.},}

\begin{document}
\title{User-Guided Domain Adaptation for Rapid Annotation from User Interactions: A Study on Pathological Liver Segmentation}

\titlerunning{User-Guided Domain Adaptation}
%
%
\author{Ashwin Raju\inst{1,2} \and Zhanghexuan Ji\inst{1,3} \and Chi Tung Cheng\inst{4} \and Jinzheng Cai\inst{1} \and Junzhou Huang\inst{2}\and Jing Xiao\inst{5} \and Le Lu\inst{1}\and ChienHung Liao\inst{4} \and Adam P. Harrison \inst{1}}

%
\authorrunning{A. Raju et al.}
%

\institute{PAII Inc., Bethesda MD, USA \and
The University of Texas at Arlington, Arlington TX, USA \and
University at Buffalo, Buffalo NY, USA \and
Chang Gung Memorial Hospital, Linkou, Taiwan, ROC \and
PingAn Technology, Shenzhen, China}
%
\maketitle              
\begin{abstract}
Mask-based annotation of medical images, especially for $3$D data, is a bottleneck in developing reliable machine learning models. Using minimal-labor \acp{PS} to guide the annotation is promising, but challenges remain on best harmonizing the mask prediction with the \acp{PS}. To address this, we propose the \ac{method} framework, which uses \ac{ADA} to model the combined distribution of \acp{PS} and mask predictions. The \acp{PS} are then used as anchors to guide and align the mask prediction. Importantly, \ac{method} can both learn from unlabelled data and also model the high-level semantic meaning behind different \acp{PS}. We test \ac{method} on annotating pathological livers using a clinically comprehensive dataset of $927$ patient studies. Using only extreme-point \acp{PS}, we achieve a mean (worst-case) performance of $96.1\%$ ($94.9\%$), compared to $93.0\%$ ($87.0\%$) for \ac{DEXTR}. Furthermore, we also show \ac{method} can retain this state-of-the-art performance even when only seeing a fraction of available \acp{PS}, demonstrating an ability for robust and reliable \ac{PS}-guided segmentation with extremely minimal labor demands. 

\keywords{Liver Segmentation \and Interactive Segmentation \and User-guided Domain Adaptation}
\end{abstract}
\acresetall
\section{Introduction}

Reliable computer-assisted segmentation of anatomical structures from medical images can allow for quantitative biomarkers for disease diagnosis, prognosis, and progression. Given the extreme labor to fully annotate data, especially for 3D volumes, a considerable body of work focuses on weakly-supervised segmentation solutions~\cite{tajbakhsh2019}. Solutions that can leverage \acp{PS}, \eg{} extreme-points, scribbles, and boundary marks, are an important such category. 



The main challenge is effectively leveraging \acp{PS} to constrain or guide the mask generation. Classic  approaches, like the \ac{RW} algorithm~\cite{Grady_2006a}, do so via propagating seed regions using intensity similarities. Later approaches add additional constraints, \eg{} based on presegmentations~\cite{Grady_2006} or learned probabilities~\cite{Harrison_2013}. With the advent of deep-learning, harmonizing mask predictions with the \acp{PS} continues to be a challenge. \Ac{DEXTR}~\cite{Maninis_2018}, which requires the user to click on the extreme boundary points of an object, is a popular and effective approach. But \ac{DEXTR} only adds the extreme point annotations as an additional channel when training the segmentor, meaning the predicted mask may not agree with the \acp{PS}.  Later work uses expectation-maximization strategies that alternate between network training and then regularization via \ac{RW}~\cite{roth2019weakly} or dense \acp{CRF}~\cite{Rajchl_2017,Wang_2018,Can_2018}. However, over and above their computational demands, the intensity-based \ac{RW} or \ac{CRF} regularization may not capture high-level semantics and any guidance on the mask predictions still remains highly indirect. DeepIGeoS~\cite{Wang_2019} offers an alternative that uses a deep-\ac{CRF} optimizer to propagate scribbles. However, DeepIGeos would only allow boundary annotations or extreme points to be treated as simple seed regions, neglecting their rich semantic meaning. 

To address these issues, we propose \ac{method}. Our new method uses \ac{ADA}  \cite{tsai2018learning} to guide mask predictions by the \acp{PS}. \ac{method}'s advantage is that it is equipped to model the high-level meaning behind different types of \acp{PS} and how they should impact the ultimate mask prediction. Importantly, the \acp{PS} are used as anchors when adapting the mask. Another advantage is that, like \ac{ADA}, \ac{method} can learn from and exploit completely unlabelled data, in addition to those accompanied by \acp{PS}. Without loss of generality, we focus on using \ac{DEXTR}-style extreme point \acp{PS} because of their intuitiveness and effectiveness~\cite{Papadopoulos_2017,roth2019weakly}. But other types, \eg{} boundary corrections, are equally possible in addition to, or instead of, extreme points.  The only constraint is that we assume a fully-supervised dataset is available in order to model the interplay between mask and \acp{PS}. But such data can originate from sources other than the target dataset, \eg{} from public data.  To the best of our knowledge, we are the first to use domain adaptation as a mechanism to drive \ac{PS}-based segmentation. 


We test \ac{method} on an extremely challenging pathological liver segmentation dataset collected directly from the \ac{PACS} of \ac{CGMH}. This dataset  comprises $927$ patient studies, all with \ac{HCC}, \ac{ICC}, metastasized, or benign lesions and is \textit{one of the most challenging datasets to date for pathological liver segmentation}. Liver volumetry analysis is a crucial pre-requisite prior to many hepatic procedures, such as liver transplantation~\cite{nakayama2006automated,taner2008donor} or resection~\cite{lodewick2016fast,lim2014ct}. Despite success with fully-automated solutions, modern deep solutions~\cite{zhang2019light,li2018h,isensee2018nnu} are often trained only on public datasets, \eg{} LiTS~\cite{bilic2019liver}, which only represents \ac{HCC} and metastasized lesions and does not fully represent patient/co-morbidity distributions. Thus, they may not generalize to all datasets, such as ours. Using \textit{only extreme-point \acp{PS}}, we achieve state-of-the-art mean (worst-case) \ac{DSC} scores of $96.1\%$ ($94.5\%$) on our dataset, compared to $93.0\%$ ($79.0\%$) and $93.1\%$ ($87.0\%$) for a strong a fully-supervised baseline and DEXTR~\cite{Maninis_2018}, respectively. We also show that \ac{method} can improve over \ac{ADA}~\cite{tsai2018learning} by $1.3\%$ \ac{DSC} and that \ac{method} can even perform robustly when only shown incomplete sets of \acp{PS}. Finally, we demonstrate predicted masks align extraordinarily well with \acp{PS}, allowing the user to interact with confidence and with minimal frustration.

\section{Methods}

\begin{figure*}[t]
	\centering
	\includegraphics[width=\linewidth]{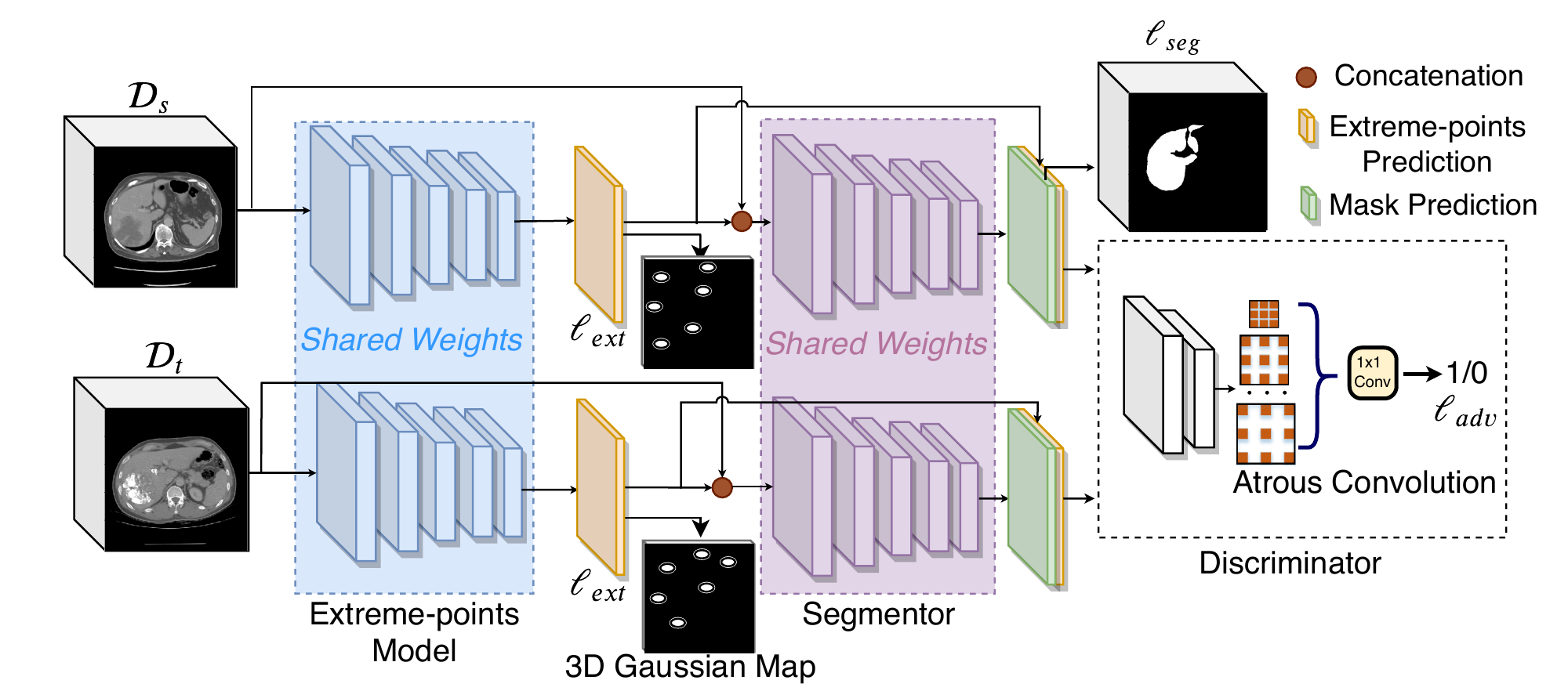}
	\caption{\textbf{Overview.} \ac{method} chains together (1) an initial \acs{FCN} that predicts the object's extreme points, and (2) a second \acs{FCN} that accepts the first's predictions to predict a mask. For source data, $\mathcal{D}_{s}$, where the mask label is present, we compute fully-supervised loss on both the extreme-point and mask predictions. For target data,  $\mathcal{D}_{t}$, we compute a fully-supervised loss when \acp{PS} are available. For all $\mathcal{D}_{t}$ volumes, whether \ac{PS}-labelled or completely unlabelled, we use \acs{ADA} to guide the mask predictions based on the extreme-point anchors. }
	\label{fig: model} \vspace{-5mm}
\end{figure*}

We aim to produce reliable mask predictions on a target dataset or deployment scenario, given only minimal \acp{PS}. More formally, we assume we are given a dataset composed of both \ac{PS}-labelled and completely unlabelled volumes, $\mathcal{D}_{t} = \{X_{i}, E_{i}\}_{i=1}^{N_{w}} \bigcup \{X_{i}\}_{i=1}^{N_{u}}$ , with $X_{i}$ and ${E}_{i}$ denoting the images and extreme points \acp{PS}, respectively. In addition, we also assume a fully-supervised source dataset with masks is also available, $\mathcal{D}_{s} = \{\mathcal{X}_{i}, Y_{i}\}_{i=1}^{N_{s}}$. As long as the masks and extreme points describe the same anatomical structure, $\mathcal{D}_{s}$ may originate from entirely different sources, \eg{} public data. The goal is to use only the extreme point \acp{PS} to robustly annotate $\mathcal{D}_{t}$. Fig.~\ref{fig: model} outlines our workflow, which uses \acf{method} to efficiently and effectively exploit the extreme-point \acp{PS}. 

\subsection{Supervised Workflow}
\label{section:s_workflow}

The backbone of \ac{method} is two 3D \acp{FCN} chained together, where the first \ac{FCN} predicts extreme-points while the second predicts a full mask. Working backward, the second \ac{FCN} acts very similarly to DEXTR~\cite{Maninis_2018}, where the latter predicts a mask given an input image along with the extreme-point \acp{PS}:
\begin{align}
    \hat{Y} = s(X, E) \mathrm{,} \label{eqn:dextr}
\end{align}
where we have used $s(.)$ to represent the segmentation \ac{FCN}. Each of the $6$ extreme points are represented by a 3D Gaussian heat map centered on the user clicks and rendered into an additional input channel, $E_{i}$. We find the method is not sensitive to the size of the Gaussian heat maps, and we use a kernel with $5$-pixel standard deviation. However, unlike DEXTR, we do not assume all training volumes come with extreme-points, as only some target volumes in $\mathcal{D}_{t}$ may have \acp{PS}. 


To loosen these restrictions, we use the first \ac{FCN} to predict extreme-point heatmaps for each volume. Following heatmap regression conventions~\cite{zhou2019bottomup}, the extreme-point \ac{FCN}, $h(.)$, outputs $6$ 3D Gaussian heatmaps, each corresponding to one extreme point. These are then summed together into one channel prior to being inputted into our segmentation \ac{FCN}:
\begin{align}
    \hat{Y} &= s(X, \hat{E}) \mathrm{,} \\
    \hat{E} &= h(X) \mathrm{,}
\end{align}
where for convenience we have skipped showing the summation of $6$ heatmaps into one channel. By relying on predictions, this allows our system to operate even with unlabelled data. Ground-truth extreme-points can be generated from any \acp{PS} performed on $\mathcal{D}_{t}$. Extreme-points for $\mathcal{D}_{s}$ can be deterministically generated from the full masks, simulating the \acp{PS}.

In terms of loss, if we use $\mathcal{D}_{e}$ to denote any input volume associated with extreme-point \acp{PS}, whether from $\mathcal{D}_{s}$ or $\mathcal{D}_{t}$, then a supervised loss can be formulated:
\begin{align}
    \mathcal{L}_{sup} &= \mathcal{L}_{seg} + \mathcal{L}_{ext} \mathrm{,} \label{eqn:sup_loss}\\ 
    \mathcal{L}_{ext} &= \dfrac{1}{N_{e}}\sum_{X,E\in\mathcal{D}_{e}} \ell_{ext}\left(h(X), E\right) \label{eqn:ext} \mathrm{,} \\
    \mathcal{L}_{seg} &= \dfrac{1}{N_{s}}\sum_{X,Y\in\mathcal{D}_{s}} \ell_{seg}\left(s(X, h(X)), Y\right) \label{eqn:seg} \mathrm{,}
\end{align}
where $N_{e}$ denotes the cardinality of $\mathcal{D}_{e}$. Following heat-map regression practices~\cite{zhou2019bottomup}, we implement $\ell_{ext}$ using mean-squared error. For $\ell_{seg}$ we use a summation of cross entropy and \ac{DSC} losses, which has experienced success in segmentation tasks~\cite{isensee2018nnu}. While we focus on extreme points for this work, other types of \acp{PS}, such as  boundary corrections, can be readily incorporated in this framework.

\subsection{User-Guided Domain Adaptation}

Similar to DEXTR~\cite{Maninis_2018}, \eqref{eqn:seg} indirectly guides mask predictions by using extreme-point heat maps as an additional input channel for the segmentor. Additionally, for \ac{PS}-labelled volumes in $\mathcal{D}_{t}$, the supervised loss in \eqref{eqn:ext} encourages the extreme-point predictions to actually match the \acp{PS}. However, the mask prediction may contradict the \acp{PS} because there is no penalty for disagreement between the two. Thus, an additional mechanism is needed to align the mask with \acp{PS}. We opt for an adversarial domain adaption approach to penalize discordant mask predictions. While image translation-based adversarial domain adaptation methods show excellent results~\cite{Hoffman_2018,Li_2019}, these are unsuited to our task because we are concerned with adapting the \emph{prediction-space} to produce a mask well-aligned with the \acp{PS}. Thus, we use \acf{ADA}~\cite{tsai2018learning}. 

More specifically, we use a discriminator, $d(.)$, to learn the distribution and interplay between masks and extreme-points. Treating samples from $\mathcal{D}_{s}$ as the ``correct'' distribution, the discriminator loss can be expressed as 
\begin{align}
\mathcal{L}_{d} = \dfrac{1}{N_{s}}\sum_{\mathcal{D}_{s}}\ell_{bce}(d(\{\hat{Y}, \hat{E}\}, \mathbf{1})) + \dfrac{1}{N_{t}}\sum_{\mathcal{D}_{t}}\ell_{bce}(d(\{\hat{Y}, \hat{E}\}, \mathbf{0})) \mathrm{,}
 \label{eqn:l_d}
\end{align}
where $\ell_{bce}$ denotes the cross-entropy loss. Importantly, to model their combined distribution, the discriminator accepts both the \ac{PS} and mask predictions. Following standard adversarial training, here gradients only flow through the discriminator. \ac{method} then attempts to fool the discriminator by predicting extreme-point/mask pairs for $\mathcal{D}_{t}$ that match $\mathcal{D}_{s}$'s distribution. More formally, an adversarial loss is set up for volumes in $\mathcal{D}_{t}$:
\begin{align}
    \mathcal{L}_{adv} = \dfrac{1}{N_{t}}\sum_{\mathcal{D}_{t}}\ell_{bce}(d(\{\hat{Y}, \hat{E}\}, \mathbf{1})) \mathrm{.} \label{eqn:l_adv}
\end{align}
Note that compared to \eqref{eqn:l_d}, the ``label'' for $\mathcal{D}_{t}$ has been switched from $\mathbf{0}$ to $\mathbf{1}$. Like standard \ac{ADA} setups, gradients do not flow through the discriminator weights in \eqref{eqn:l_adv}. Importantly, \emph{gradients also do not flow through the extreme-point predictions when the \acp{PS} are present}. Consequently, when \acp{PS} are available, extreme-point predictions are only influenced by the supervised loss in \eqref{eqn:ext} to match the \acp{PS}. While there is no strict guarantee, our results demonstrate almost perfect matching. Thus, the extreme-point predictions act as anchors, while the adversarial loss in \eqref{eqn:l_adv} guides the mask predictions to properly align with the \acp{PS}. This alignment is more than simply making mask predictions agree with the \acp{PS}, as by modelling their interplay, \ac{ADA} also guides mask regions far away from \acp{PS}. Finally, the use of \ac{ADA} provides another important benefit, as completely unlabelled volumes in $\mathcal{D}_{t}$ can seamlessly contribute to the learning process in \eqref{eqn:l_d} and \eqref{eqn:l_adv}. In fact, \ac{method} can be seen as integrating domain adaptation learning processes~\cite{tsai2018learning} in addition to DEXTR-style guidance~\cite{Maninis_2018} from \acp{PS}. Thus, the overall training objective for \ac{method} is to minimize the following total loss:
\begin{align}
   \mathcal{L} = \mathcal{L}_{sup} + \lambda_{adv}\mathcal{L}_{adv} \mathrm{,} \label{eqn:total}
\end{align}
where we have purposely kept loss weighting to only the adversarial component to reduce hyper-parameter tuning.  


\section{Experiments}
\begin{figure}[t]
\center
      \includegraphics[width=\linewidth]{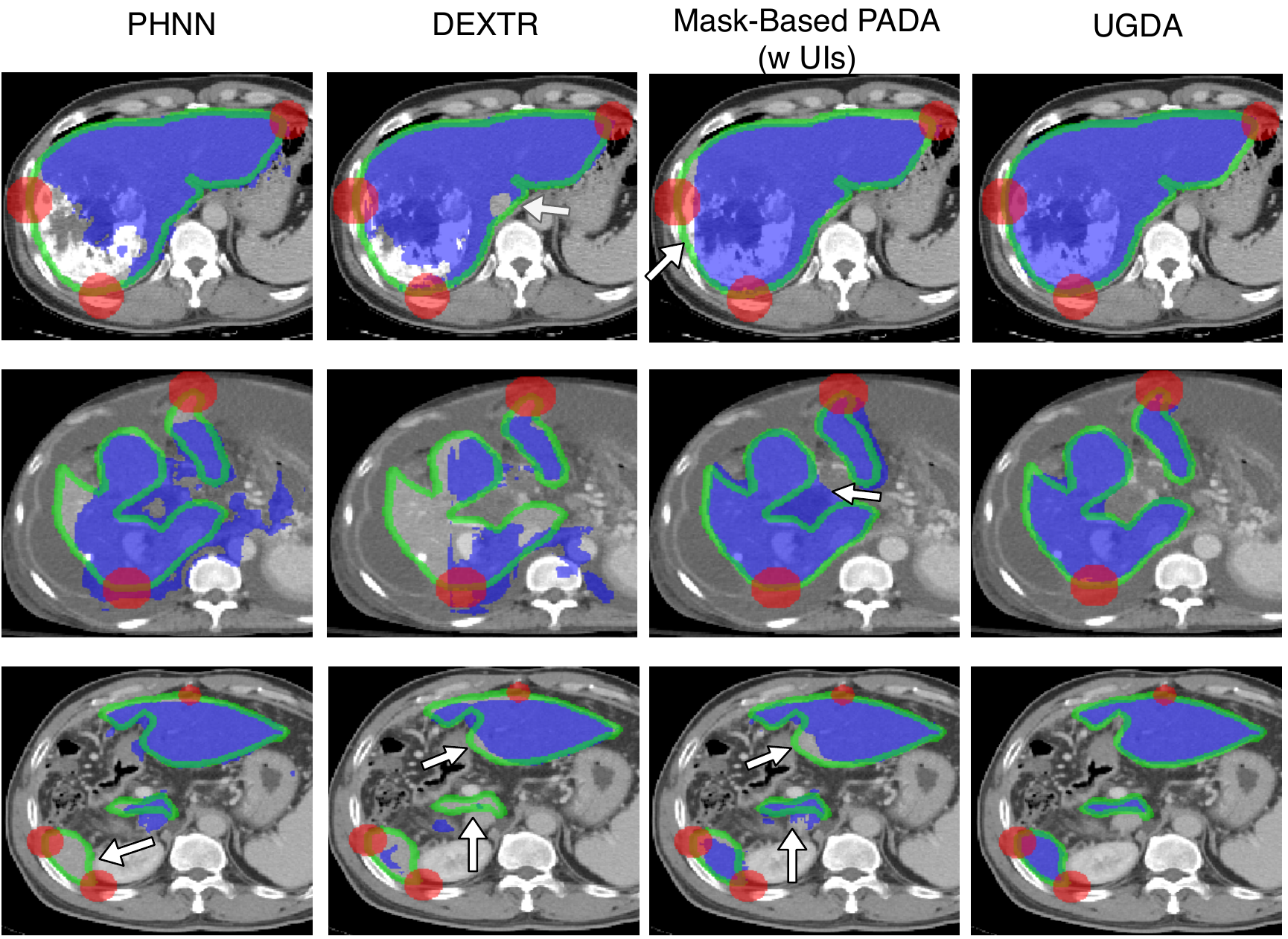}
     \caption{\textbf{Qualitative results.} Liver mask ground truth and predictions are rendered in green contour and blue mask, respectively, and Guassian heatmaps centered on the extreme point \acp{PS} are shown in red. 
     As can be seen, \ac{method} can much better align the masks with the \acp{PS}. 
     Arrows highlight selected baseline prediction errors that \ac{method} corrects. (Best viewed in color)}
     \label{fig:qualitative} \vspace{-3mm}
\end{figure}
\textbf{Dataset description.} We test \ac{method} on segmenting pathological livers, using a target dataset of $927$ venous-phase \ac{CT} studies from the \ac{PACS} of \ac{CGMH}.  The only selection criterion was of patients with biospied or resecuted liver lesions, with \ac{CT} scans taken within one month before the procedure. Patients directly reflect clinical distributions and represent \ac{ICC}, \ac{HCC}, benign or metastasized lesions, along with co-occuring maladies, such as liver fibrosis, splenomegaly, or embolized lesions. This serves as $\mathcal{D}_{t}$. From these, we selected $47$ and $100$ studies as validation and test sets, respectively, and delineated the patient livers. These $147$ \acp{CT} are called \emph{evaluation volumes}. We annotated the remainder using only extreme point \acp{PS}.  For $\mathcal{D}_{s}$ we collected $235$ fully-labelled venous-phase \ac{CT} studies from public datasets~\cite{bilic2019liver,gibson_eli_2018_1169361,4781564,chaos}, which, unlike $\mathcal{D}_{t}$, comprises both healthy and pathological livers and only represents \ac{HCC} and metastasized tumors. Corresponding extreme-point ``\acp{PS}'' were generated from the full masks following~\cite{Maninis_2018}. For internal validation, we also split $\mathcal{D}_{s}$   into $70\%$, $20\%$, and $10\%$ for training, testing and validation, respectively.


\textbf{Implementation Details.}  For \ac{method}'s two \ac{FCN} architectures, we use a $3$D version of the deeply-supervised \ac{PHNN}~\cite{harrison2017progressive}, which provides an efficient and decoder-free pipeline. As backbone, we use a $3$D generalization of VGG-16~\cite{Simonyan15}.  We first train a fully-supervised baseline on $\mathcal{D}_{s}$ using \eqref{eqn:sup_loss}, then we finetune after convergence using \eqref{eqn:total}. This dual-PHNN baseline is very strong on our public data, achieving a \ac{DSC} score of $96.9\%$ on the $\mathcal{D}_{s}$ test set. For discriminator, we use a $3$D version of a popular architecture~\cite{Lv_2018_ECCV_Workshops} using atrous convolution, which has proved a useful discriminator for liver masks~\cite{raju2020co}. We do not use the multi-level discriminator variant proposed by Tsai \etal{}~\cite{tsai2018learning}, as its added complexity does not seem to result in noticeable improvements for liver-based \ac{ADA}~\cite{raju2020co}. Specific details on the choice of hyper parameters are listed in the supplementary material.

\textbf{Evaluation protocols.} We evaluate \ac{method} on how well it can annotate $\mathcal{D}_{t}$ using only extreme-point \acp{PS}. To do this, we include the evaluation volumes and their extreme point \acp{PS} within the training procedure, \textit{but hide their masks}. For evaluation, we measure \ac{DSC} scores and also the \ac{MXA}. The latter measures the average distance between all six of a predicted 
\emph{mask's} extreme-points vs. the ground-truth extreme points. This directly measures how well the model can produce a mask prediction that actually matches the extreme-point \acp{PS}.  

\textbf{Comparisons.} We test against the user-interactive DEXTR~\cite{Maninis_2018} using the same \ac{PHNN} backbone. DEXTR can only be trained on $\mathcal{D}_{s}$ because it requires fully-labelled training data, but during inference it sees $\mathcal{D}_{t}$'s extreme-point \acp{PS}. The DEXTR authors claim their approach can generalize well to new data. We also test against two  mask-based \ac{ADA}~\cite{tsai2018learning} variants. The first variant, called ``mask-based \ac{ADA} (no \acp{PS})'', matches published practices~\cite{tsai2018learning} and just uses a single network (\ac{PHNN} in our case) to directly predict masks, using a discriminator to penalize ill-behaving mask predictions on unlabeled data. It does not incorporate \acp{PS}. The second variant, simply called ``mask-based \ac{ADA} (w \acp{PS})'', is almost identical to \ac{method}, meaning the mask and extreme-point predictions are still trained with the supervised losses of \eqref{eqn:seg} and \eqref{eqn:ext}, respectively, but the discriminator in \eqref{eqn:l_d} and \eqref{eqn:l_adv} only models mask predictions without seeing the extreme-point predictions. Comparing this variant to \ac{method} reveals the impact of using extreme-points as anchors to guide the domain adaptation of the masks.   

Finally, we also compare against \ac{method} variants trained with only a fraction of the \acp{PS} available in $\mathcal{D}_{t}$, with the remainder being left unlabelled. This reveals how well \ac{method} can operate in scenarios when only a fraction of target volumes have \ac{PS}-labels. When doing so, we remove the same percentage of \ac{PS}-labels from both non-evaluation and evaluation volumes.

\begin{figure}[t]
\center
      \includegraphics[width=.5\linewidth]{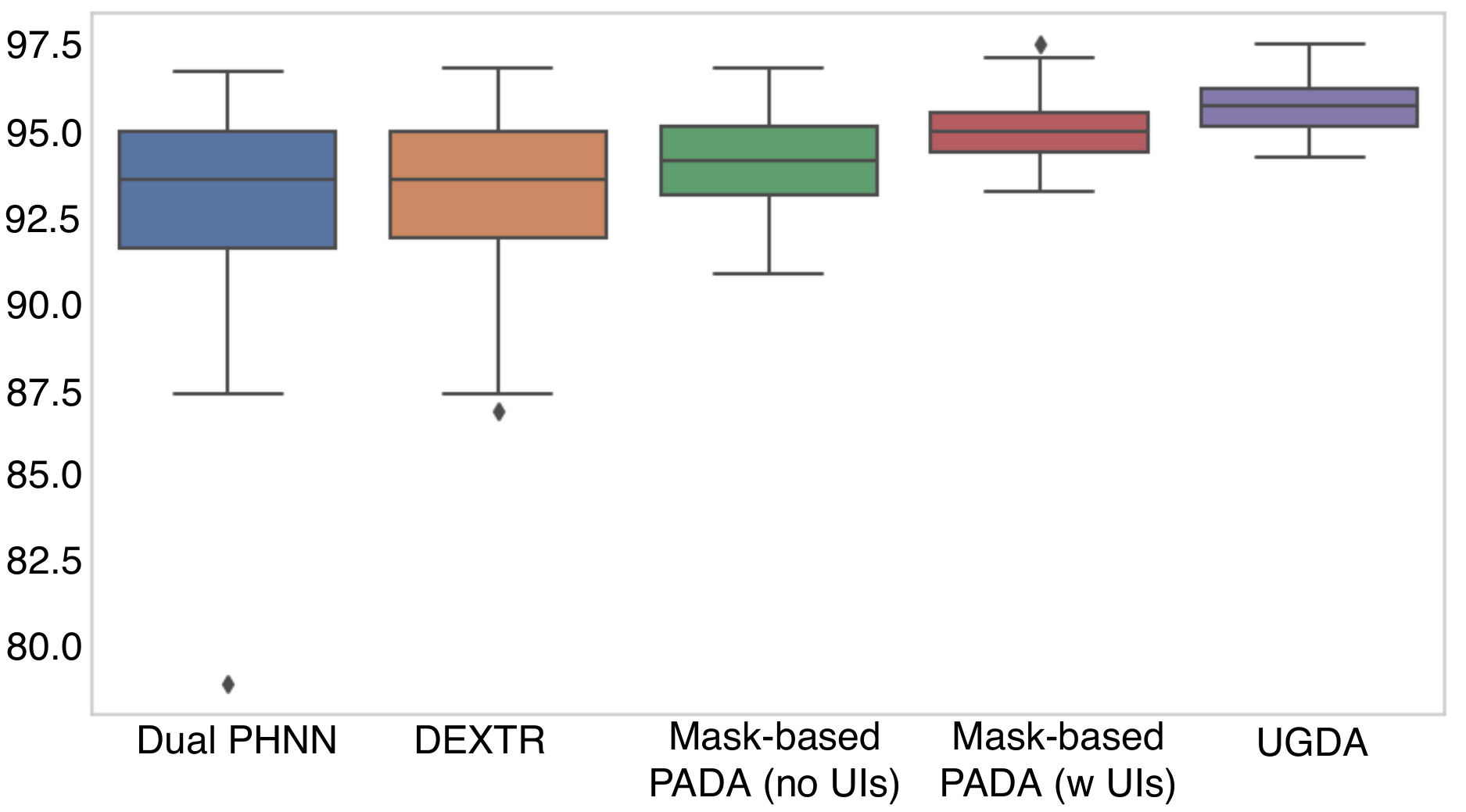}
      \caption{\textbf{Box and whisker plot} of pathological liver segmentation DSC scores across the $100$ $\mathcal{D}_{t}$ test set volumes.}
      \label{fig:mask_bw_interactive}\vspace{-5mm}
\end{figure}

\textbf{Results.} Tab.~\ref{tab:interactive_results} outlines the performance of all variants in annotating $\mathcal{D}_{t}$. As can be seen, compared to its performance on $\mathcal{D}_{s}$, the fully-supervised dual \ac{PHNN}'s performance drops from $96.9\%$ to $93.0\%$ due to the major differences between public liver datasets and our \ac{PACS}-based clinical target dataset. This suggests that other strategies are required, \eg{} exploiting minimal-labor \acp{PS}.
\begin{table}[t]
\caption{\ac{DSC} and \ac{MXA} mean and standard deviation scores. In parentheses are the fraction of \ac{PS}-labelled $\mathcal{D}_{t}$ volumes  used for training (or inference for DEXTR).}	
	\begin{center}	
	\footnotesize
		\begin{tabular}{ |p{5.1cm}|>{\centering}p{2.2cm}|>{\centering}p{2.2cm}|>{\centering\arraybackslash}p{2.2cm}|}\hline
			{\bfseries Model} & {\bfseries \% \acp{PS}} &  {\bfseries Mean DSC}  & {\bfseries Mean \acs{MXA} (mm)} \\ \hline
			Dual PHNN~\cite{harrison2017progressive} & n/a & $93.0 \pm 3.2 $ &  $4.3 \pm 1.2  $ \\ \hline
			DEXTR~\cite{Maninis_2018} & $100\%$   & $ 93.1 \pm 2.4  $ & $3.9 \pm 1.2 $ \\ \hline
			Mask-based PADA (no \acp{PS})~\cite{tsai2018learning} & $0\%$ & $94.8 \pm 1.8$ &  $3.4 \pm 1.6$ \\ \hline
		    Mask-based PADA (w \acp{PS})~\cite{tsai2018learning} & $100\%$ & $95.5 \pm 1.0 $ & $2.5 \pm 1.0 $  \\ \midrule
		    \ac{method} & $25\%$  & $95.8 \pm 0.8$ & $1.7 \pm 0.8 $ \\ \hline
		    \ac{method} & $50\%$  & $96.0 \pm 0.9$ & $1.4 \pm 0.9$ \\ \hline
			\ac{method} &  $100\%$  & $\mathbf{96.1 \pm 0.8}$ & $\mathbf{1.1 \pm 0.9}$ \\ \hline
			
		\end{tabular}
	\end{center}
\label{tab:interactive_results} \vspace{-4mm}
\end{table}
As the table also demonstrates, by exploiting \acp{PS}, DEXTR can significantly boost the \ac{MXA}, but its lackluster \ac{DSC} scores suggest that the resulting masks, while aligning better with the extreme-points, still do not properly capture the liver extent. On the other hand, both mask-based \ac{ADA} variants perform better, indicating that modelling mask distributions on top of \ac{DEXTR}-style \ac{PS} guidance can more robustly annotate $\mathcal{D}_{t}$. Finally, \ac{method} performs best, demonstrating that modelling the \emph{interplay} between the \acp{PS} and mask predictions can boost performance even further. Importantly, \ac{method}'s \ac{MXA} is extremely good,  which shows that the mask predictions well match the \acp{PS}. These mean scores are bolstered by Fig.~\ref{fig:mask_bw_interactive}'s box-and-whisker plots, which demonstrate that \ac{method} provides important boosts in reliability, with an extremely robust worst-case performance of $94.9\%$ \ac{DSC}, compared to $93.2\%$ for the mask-based \ac{ADA} (w \acp{PS}) variant.
Viewing the qualitative examples in Fig.~\ref{fig:qualitative} reinforces these quantitative improvements. In particular, \ac{method} is able to ensure that masks both agree with extreme-points and provide robust predictions away from the \acp{PS}. 

Finally, as Tab.~\ref{tab:interactive_results} also demonstrates, \ac{method} can perform almost as well when only a fraction of $\mathcal{D}_{t}$ is \ac{PS}-labelled, outperforming both \ac{DEXTR} and the mask-based \ac{ADA}, both of which see all $100\%$ of the \acp{PS}. These results indicate that \ac{method} can operate well even in scenarios with \emph{extremely minimal} \ac{PS} annotation. providing further evidence of its very high versatility.

\section{Conclusion}

We presented \acf{method}, an effective approach to use \acp{PS} to rapidly annotate 3D medical volumes. 
We tested \ac{method} on arguably the most challenging and comprehensive pathological liver segmentation dataset to date, demonstrating that by only using extreme-point \acp{PS} we can achieve \ac{DSC} scores of $96.1\%$, outperforming both \ac{DEXTR}~\cite{Maninis_2018} and conventional mask-based \ac{ADA}~\cite{tsai2018learning}. Future work includes testing on other anatomical structures, incorporating additional types of \acp{PS}, and adapting \ac{method} for real-time interaction.

\begin{center}
    {\large\bfseries Supplementary material}

\end{center}
\section*{Hyper parameters}
We set $\lambda_{adv}$ to $0.0001$. The dual-\ac{PHNN} learning rates was set to $0.003$ and reduced by a factor of $0.1$ when validation \acp{DSC} do not improve after $15$ epochs, whereas the discriminator has a constant learning rate of $0.0003$. We use the Adam optimizer~\cite{Kingma_2015} for both. 

\begin{table}[ht]
\caption{Some of the pertinent details we followed for all our experiments.}	
	\begin{center} 	
	\footnotesize
		\begin{tabular}{ |c|c|}\hline
		Input volume resolution & $256 \times 256 \times 48$  \\  \hline
		Standard deviation for $3$D Gaussian map & 5 \\ \hline
		GPU system & Quadro RTX 8000  \\ \hline
		Cuda version & 10.2  \\ \hline
		Number of GPUs and memory& $4\times 48GB$ \\ \hline
		Deep learning framework and version & pytorch 1.4.0 \\ \hline

		\end{tabular}
	\end{center}
\label{tab:interactive_results} \vspace{-4mm}
\end{table}
\newpage
\bibliographystyle{splncs04}
\bibliography{sample}
\end{document}